\documentclass[letterpaper, 10 pt, conference]{ieeeconf}  

\IEEEoverridecommandlockouts                              
\usepackage{graphics} 
\usepackage{mathtools}
\usepackage{bm}
\usepackage{epsfig} 
\usepackage{mathptmx} 
\usepackage{times} 
\usepackage{amsmath} 
\usepackage{amssymb}  
\usepackage{nccmath}
\usepackage{hyperref}
\usepackage{ragged2e}

\title{\LARGE \bf
OmBURo: A Novel Unicycle Robot with Active Omnidirectional Wheel
}

\author{Junjie Shen$^1$ and Dennis Hong$^1$
\thanks{
$^1$J. Shen and D. Hong are with the Department of Mechanical and Aerospace Engineering, University of California, Los Angeles, CA 90095 USA (e-mail: \href{mailto:junjieshen@ucla.edu}{junjieshen@ucla.edu}; \href{mailto:dennishong@ucla.edu}{dennishong@ucla.edu}).
}
}

\begin{document}

\maketitle
\thispagestyle{empty}
\pagestyle{empty}

\begin{abstract}

A mobility mechanism for robots to be used in tight spaces shared with people requires it to have a small footprint, to move omnidirectionally, as well as to be highly maneuverable. However, currently there exist few such mobility mechanisms that satisfy all these conditions well. Here we introduce Omnidirectional Balancing Unicycle Robot (OmBURo), a novel unicycle robot with active omnidirectional wheel. The effect is that the unicycle robot can drive in both longitudinal and lateral directions simultaneously. Thus, it can dynamically balance itself based on the principle of dual-axis wheeled inverted pendulum. This letter discloses the early development of this novel unicycle robot involving the overall design, modeling, and control, as well as presents some preliminary results including station keeping and path following. With its very compact structure and agile mobility, it might be the ideal locomotion mechanism for robots to be used in human environments in the future.

\end{abstract}

\section{Introduction}

On flat surfaces, wheeled locomotion is the most widely used mobility mechanism for robots. It has high energy efficiency since an ideal rolling wheel requires no further energy input to sustain its motion. This is in contrast to legged robots which suffer energy loss at heal-strike. For simplicity, most mobile robots have three or more wheels to achieve static stability. However, they can easily become dynamically unstable if the center of mass is too high, or the base of support is too narrow, or the acceleration is too large, or simply the slope is too steep \cite{ballbot}. As a result, the performance is significantly limited for statically-stable wheeled mobile robots. 

On the other hand, many interesting mobile robots using only one or two wheels have been created over the past three decades \cite{Schoonwinkel:1988:DTC:914106, MIT, Tsukuba}. These robots are dynamically stabilized by closed-loop control. That is, they are able to keep their balance autonomously without the need of static stability from the structure configuration. More interestingly, they utilize inertial for locomotion. In order to move in some direction, there is an initial retrograde wheel movement causing the body to lean towards the goal direction due to inertia, and then the wheels reverse and speed up. As performance limits of mobile robots are pushed, dynamic effects will increasingly come into play to enhance mobility \cite{dynamic, Xu}. This becomes extremely true when it comes to robots operating in human environments. The robots are desired to be dynamically stable for better maneuverability, capable of omnidirectional motion, slim enough for moving in a crowd, and yet simple for the sake of safety as well as maintenance.
\begin{figure}[!t]
\centering
\includegraphics[width=4.3cm]{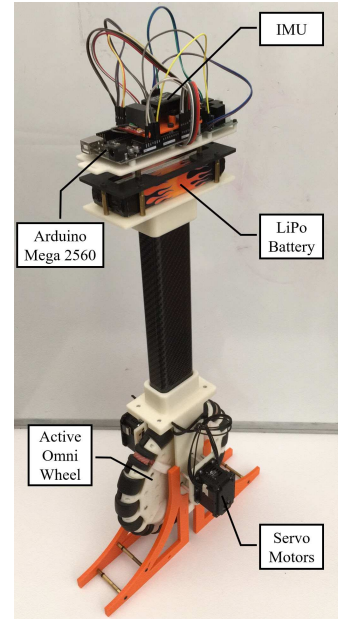}
\caption{OmBURo prototype}
\label{fig: unicycle robot}
\end{figure}

Some typical examples of dynamically-stable wheeled mobile robots have emerged to this day. Segway type robots \cite{Tsukuba, segway1, segway2, segway3, segway4} with two wheels are able to balance in one direction but cannot immediately move in the lateral direction without first re-orienting the drive system, and thus it cannot maneuver in highly dynamic environments. Besides, it might fall over on an inclined plane if not aligned with its moving direction. The cross section is also relatively large which can create problems in narrow spaces.

A conventional unicycle robot uses a single wheel for longitudinal balancing and another reaction wheel for lateral balancing, which however results in a quite complicated and bulky mechanical structure \cite{Schoonwinkel:1988:DTC:914106, MIT, uni1, uni2, uni3, uni4, uni5, uni6, uni7, uni8}. It cannot move omnidirectionally either.

A robot that uses a ball instead of a wheel (a.k.a. Ballbot \cite{ballbot, ballbot2, ballbot3}) can move in any arbitrary direction, but its ball drive mechanism makes it unsuitable for ill-conditioned terrains. Essentially, the ball is driven by some rollers and this only works when there is sufficient friction between them. If the ground is wet or dirty, the ball surface can easily be influenced and the coefficient of friction will thus drop dramatically \cite{ballwheel, cof}. The transmission may even fail in some extreme scenarios, e.g., slipping or stuck by some small particles from the ground.


\begin{figure}[!t]
\begin{minipage}[t]{0.5\linewidth}
\centering
\includegraphics[height=5.5cm]{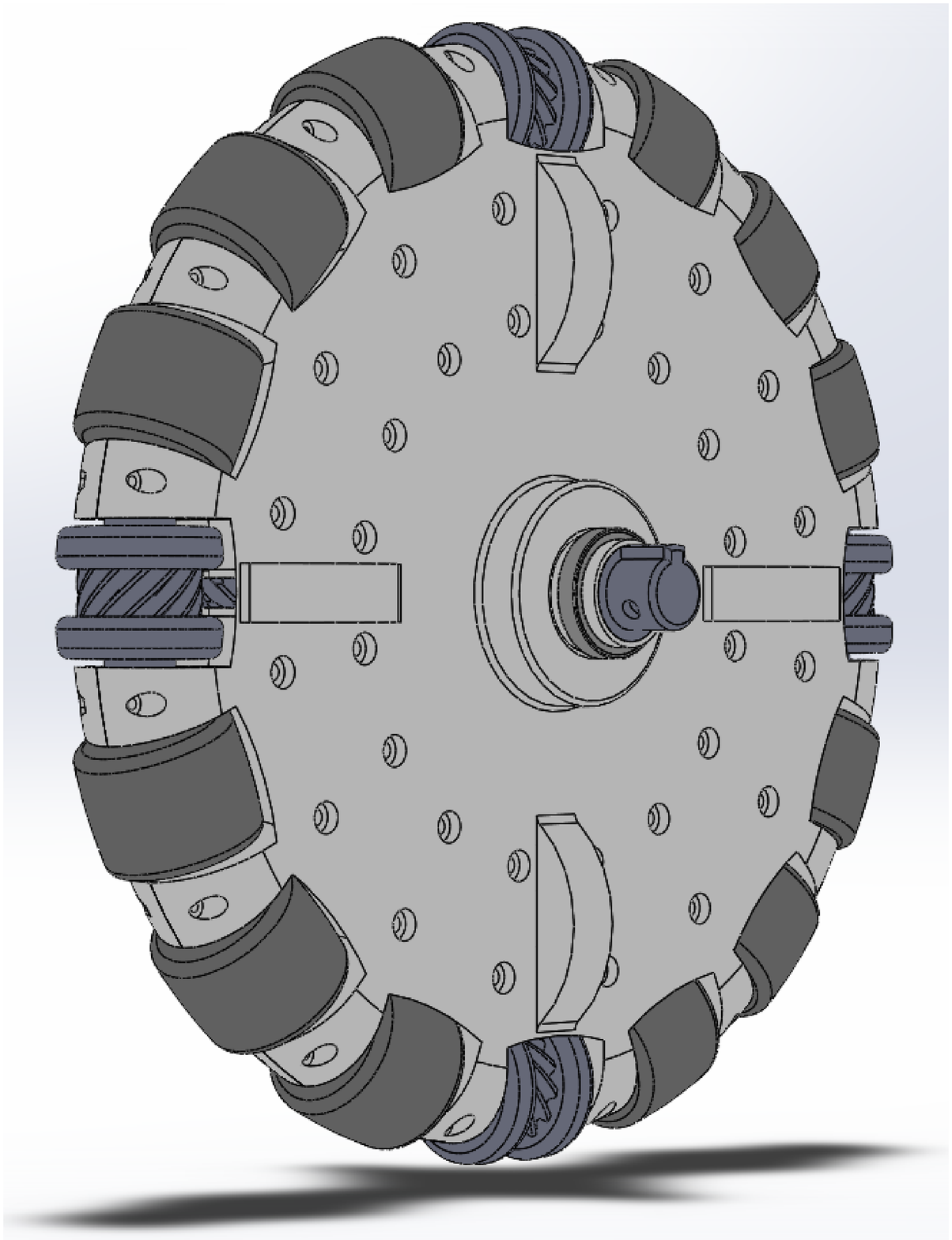}
\end{minipage}%
\begin{minipage}[t]{0.5\linewidth}
\centering
\includegraphics[height=5.5cm]{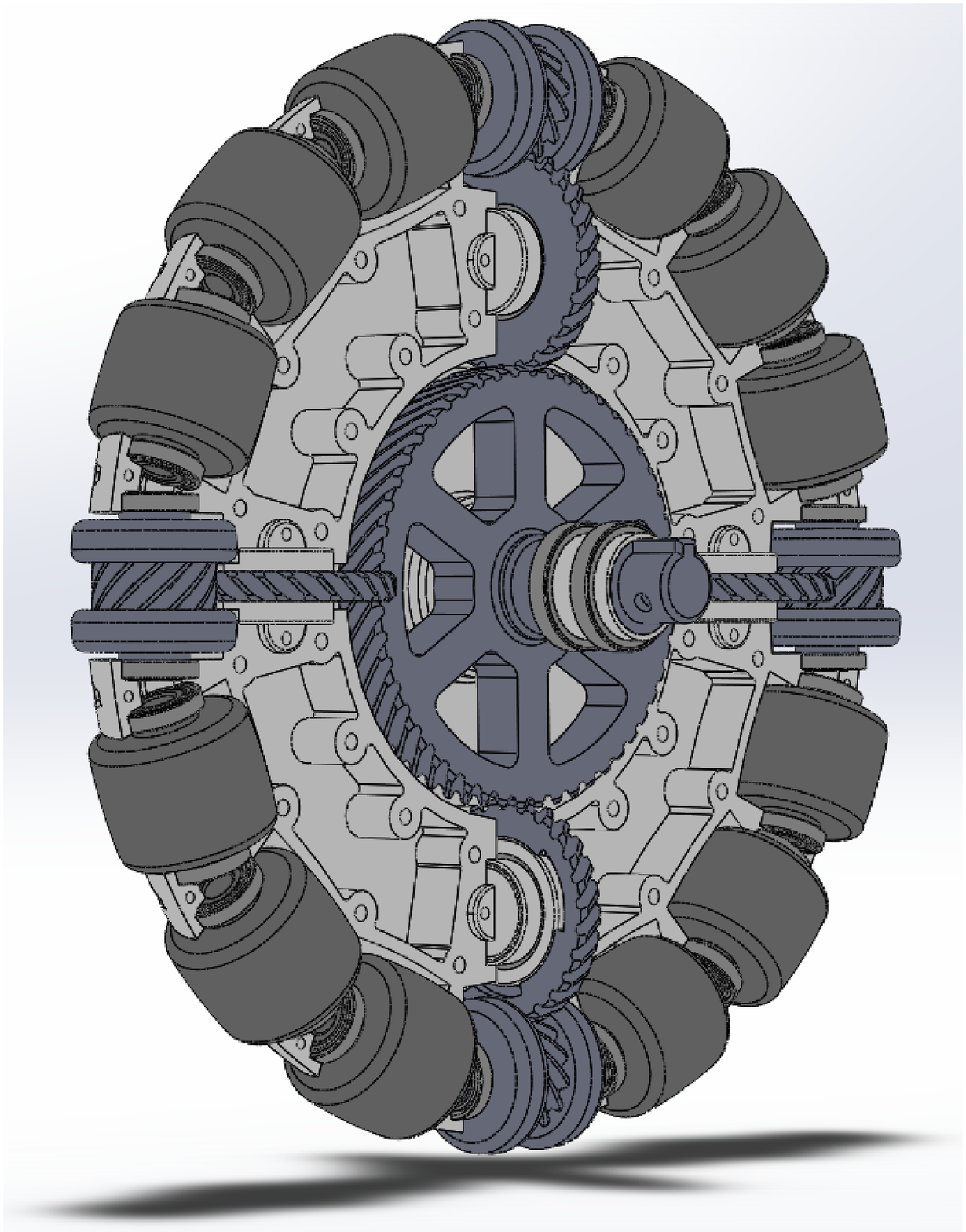}
\end{minipage}
\caption{Active omnidirectional wheel with/without housing enclosure}
\label{fig: active omniwheel}
\end{figure}

Last but not least, Honda U3-X \cite{u3x} is a mono-wheel personal mobility device unveiled in 2009. It's equipped with Honda Omni Traction Drive System, which is known as the world's first wheel structure that enables movement in all directions. U3-X is highly compact and easy to maintain due to its unique roller-on-tire friction drive transmission. However, its capabilities are consequently limited. Similar to Ballbot, it is desired to operate only on a clean and dry surface. Besides, it suffers rapid tire wear \cite{tirewear}. A normal tire makes contact with the ground once every revolution. On U3-X, the tire now has many more points of contact with the driving rollers which operate at a pretty high speed. Furthermore, the system needs to overcome velocity difference at these contact points, which in turn accelerates tire wear and causes energy loss as heat \cite{heat}.

Inspired by the previous work, we have been developing OmBURo: a novel unicycle robot with active omnidirectional wheel. The mobility mechanism is desired to be as elegant as friction drive while avoiding its intrinsic problems, e.g., terrain limitation, tire wear, energy loss. The rest of this letter is organized in the following manner: Section II describes the physical system; mathematical modeling is derived in Section III; locomotion control strategy is elaborated in Section IV; experimental results are shown in Section V; and Section VI summarizes our contributions and future work. 





\section{System Description}

This section describes the physical system of the prototype as shown in Fig. \ref{fig: unicycle robot}. The major components are the active omnidirectional wheel, actuators, and sensors.


\subsection{Active Omnidirectional Wheel}

Omnidirectional wheels are wheels with small free rollers around the circumference which are perpendicular to the spinning direction. The effect is that omni wheels can be driven in both longitudinal and lateral directions without changing the orientation of the main wheel. However, since the
rollers are passive, several omni wheels have to work in conjunction with each other to perform the desired driving maneuvers such as lateral side to side movement. 

Recent studies have been focused on how to actuate the peripheral rollers in a conventional omni wheel (thus called \textit{active omnidirectional wheel}) so that a single wheel unit can actively move in any arbitrary direction. \cite{omniwheel2} first proposed a differential gear mechanism (in total 3 gear pairs with 1:24 gear ratio from input shaft to roller) to drive each roller. In order to accommodate as many rollers as possible for a smoother circumference, the output gear of the differential has to have a small radius, which in turn sacrifices output torque dramatically at the roller. In the end, it would be difficult to satisfy design requirements for highly dynamic locomotion. To increase output torque, \cite{omniwheel1} inserted a long gear train between the differential and the roller (5 gear pairs with 1:8 gear ratio). However, this deteriorates energy loss because more gear pairs result in more gear friction to overcome \cite{gear_friction}. Besides, not all the rollers are directly driven via the gear mechanism but connected to the driven ones via universal joint, which is not a constant-velocity joint, i.e., the rollers have different rotating speeds in operation. \cite{omniwheel3} used double universal joint to solve the problem but ended up with a quite large gap between the rollers, causing lethal vibration issues when rolling.

\begin{figure}[!t]
\centering
\includegraphics[width=5.4cm]{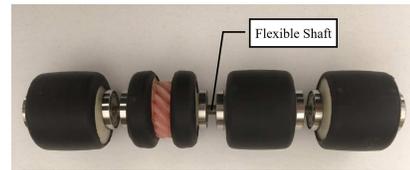}
\caption{Every four rollers are attached to one flexible shaft. One of them is driven via the helical gears.}
\label{fig: flexible shaft}
\end{figure}

Fig. \ref{fig: active omniwheel} shows the mechanical structure of the proposed active omnidirectional wheel, aiming to improve the existing works for better implementation on a unicycle robot. Among the sixteen rollers, only four of them are directly driven via the helical gears. The gear mechanism is not used for all the rollers mainly to reduce energy loss from gear friction as well as weight and inertia for better maneuverability. To drive the rest rollers, flexible shaft is used as shown in Fig. \ref{fig: flexible shaft}. It is a rotating coil which is flexible in bending but still has high torsional stiffness, so that constant velocity is guaranteed among the rollers. Compare to \cite{omniwheel2, omniwheel1, omniwheel3}, it has larger output torque due to a higher gear ratio (1:4) and better energy efficiency due to a shorter gear train (2 gear pairs); All the rollers are synchronized while the gap between them is smaller, leading to a desired smoother rolling motion.

\subsection{Actuators}

The prototype uses two Dynamixel MX-64 servo motors as actuators. One is directly driving the main wheel in the longitudinal direction while the other is actuating the rollers laterally through the helical gears and flexible shaft.

\subsection{Sensors}

A LORD Microstrain 3DM-GX4-25 Inertial Measurement Unit (IMU) provides Kalman-filtered Euler angles and rates of the unicycle body with respect to gravity. Additionally, each servo motor is equipped with a 4096 PPR encoder (i.e., 0.088$^{\circ}$ resolution) detecting motor shaft angular position and velocity. Full-state feedback control can thus be achieved.

\subsection{Others}

The prototype is designed to be completely self-contained, powered by a 14.4 V 3300 mAh LiPo battery and controlled by an Arduino Mega 2560 board at 125 Hz. The Arduino communicates with the servo motors via an RS-485 interface while the IMU via an RS-232 interface.

\section{Modeling}

This section derives the mathematical modeling of kinematics and dynamics for OmBURo.

\subsection{Kinematics}

\begin{figure}[!t]
    \centering
	\includegraphics[height=6cm]{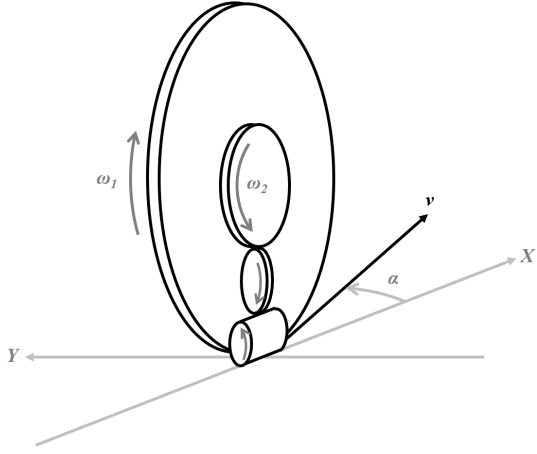}
	\caption {Velocity kinematics diagram}
	\label{fig: velocity kinematics}
\end{figure}

Fig. \ref{fig: velocity kinematics} shows the velocity kinematics diagram of OmBURo. Note that the kinematic model is derived assuming it is always at the nominal upright configuration.

Given the motor speeds $\omega_1$ and $\omega_2$, the velocity $v$ of OmBURo is determined to be
\begin{align*}
\underbrace{\left[\begin{array}{cc}
v_x\\
v_y
\end{array}\right]}_v=\left[\begin{array}{cc}
\omega_1R\\
n\left(\omega_1+\omega_2\right)r
\end{array}\right]=\underbrace{\left[\begin{array}{cc}
R&0\\
nr&nr
\end{array}\right]}_{\bm{J}}\left[\begin{array}{cc}
\omega_1\\
\omega_2
\end{array}\right], \tag{1}\label{eqn1}
\end{align*}
where $n$ is the inverse of the gear ratio. The radius of the main wheel is denoted as $R$. The radius of the roller is denoted as $r$. Since the Jacobian matrix $\bm{J}$ is nonsingular, we can inversely calculate the motor speeds based on the desired OmBURo moving velocity. Table \ref{tab: velocity kinematics} summarizes the relationship between the motor speeds and OmBURo moving direction $\alpha$.

\begin{table}[ht]
\caption{OmBURo Moving Direction}
\begin{center}
\begin{tabular}{c|c}
\hline
\bf{Motor Speeds}&\bf{Moving Direction}\\\hline
$\omega_1>0,\quad\omega_1+\omega_2=0$&$\alpha=0$\\\hline
$\omega_1>0,\quad\omega_1+\omega_2>0$&$0<\alpha<\pi/2$\\\hline
$\omega_1=0,\quad\omega_1+\omega_2>0$&$\alpha=\pi/2$\\\hline
$\omega_1<0,\quad\omega_1+\omega_2>0$&$\pi/2<\alpha<\pi$\\\hline
$\omega_1<0,\quad\omega_1+\omega_2=0$&$\alpha=\pi$\\\hline
$\omega_1<0,\quad\omega_1+\omega_2<0$&$\pi<\alpha<3\pi/2$\\\hline
$\omega_1=0,\quad\omega_1+\omega_2<0$&$\alpha=3\pi/2$\\\hline
$\omega_1>0,\quad\omega_1+\omega_2<0$&$3\pi/2<\alpha<2\pi$\\\hline
\end{tabular}
\label{tab: velocity kinematics}
\end{center}
\end{table}

To derive the position kinematics, we can simply integrate both sides of (\ref{eqn1}) over time and end up with
\begin{align*}
\left[\begin{array}{cc}
p_x\\
p_y
\end{array}\right]=\bm{J}\left[\begin{array}{cc}
\beta_1\\
\beta_2
\end{array}\right], \tag{2}
\end{align*}
where $p_x$ and $p_y$ indicate the position of OmBURo while $\beta_1$ and $\beta_2$ indicate the motor rotation angles.

\subsection{Dynamics}


\begin{figure}[!t]
\centerline{\includegraphics[width=8.5cm]{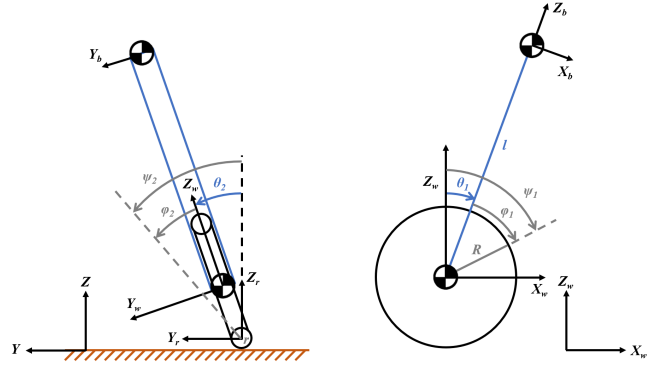}}
\caption{OmBURo model}
\label{fig: model}
\end{figure}

Some major assumptions are established in the first place: all the components are rigid bodies; the ground is even and hard; the wheel is always in contact with the ground without slipping; and yaw dynamics is negligible. 

Fig. \ref{fig: model} shows the model of OmBURo in detail. The inertial frame is denoted as $XYZ$. The roller frame is denoted as $X_rY_rZ_r$ corresponding to the inertial frame. The wheel frame is denoted as $X_wY_wZ_w$ with the $X_w-$axis always parallel to the $X_r-$axis. The body frame is denoted as $X_bY_bZ_b$ with the $Y_b-$axis always parallel to the $Y_w-$axis. $\theta_1$ is the body angle from $Z_w$ to $Z_b$ along $Y_w$. $\theta_2$ is the body angle from $Z_r$ to $Z_w$ along $X_r$. $\psi_1$ is the rotation angle of the wheel along $Y_w$. $\psi_2$ is the rotation angle of the roller along $X_r$. $\varphi_1$ is the relative rotation angle between the body and the wheel along $Y_w$. $\varphi_2$ is the relative rotation angle between the body and the roller along $X_r$. The distance between the wheel frame and the body frame is denoted as $l$.

Lagrange formulation is used to derive the equations of motion which take the form:
\begin{align*}
\bm{M}(\bm{q})\bm{\ddot{q}}+\bm{C}(\bm{q},\bm{\dot{q}})+\bm{F}(\bm{\dot{q}})+\bm{G}(\bm{q})=\bm{W}(\bm{u}),\tag{3}\label{eqn3}
\end{align*}
where $\bm{q}=\left[\theta_1, \theta_2, \varphi_1, \varphi_2\right]^T$ is the vector of generalized coordinates and $\bm{u}=\left[u_1, u_2\right]^T$ is the vector of generalized forces applied at
$\bm{\varphi}=\left[\varphi_1,  {\varphi}_2\right]^T$, $\bm{M}(\bm{q})$ is the inertia matrix, $\bm{C}(\bm{q},\bm{\dot{q}})$ represents the vector of Coriolis and centrifugal forces, $\bm{F}(\bm{\dot{q}})$ represents the vector of viscous friction forces, $\bm{G}(\bm{q})$ stands for the vector of gravitational forces, and $\bm{W}(\bm{u})$ stands for the vector of control inputs. The details of derivation are shown in Appendix \ref{dynamics}.

Note that the two relative rotation angles $\varphi_1$ and $\varphi_2$ are not involved in the equations of motion since they only determine the position of OmBURo due to the non-slipping assumption. And the position would not influence its dynamics based on the uniform-terrain assumption.

\newpage
\section{Control}


This section discusses the development of locomotion control strategy for OmBURo. 

\subsection{State-space Realization}

Since the equations of motion are highly nonlinear and complicated while we always expect OmBURo to operate in the neighborhood of the nominal upright configuration $(\theta_1=\theta_2\approx0\textrm{, }\dot{\theta}_1=\dot{\theta}_2\approx0)$, it's reasonable to linearize (\ref{eqn3}) at this configuration, which yields
\begin{align*}
&a_1\ddot{\theta}_1+a_2\ddot{\varphi}_1+\mu_g\dot{\theta}_1+\mu_g\dot{\varphi}_1+a_3\theta_1=0,\tag{4}\label{eqn4}\\
&a_4\ddot{\theta}_2+a_5\ddot{\varphi}_2+\mu_g\dot{\theta}_2+\mu_g\dot{\varphi}_2+a_6\theta_2=0,\tag{5}\label{eqn5}\\
&a_2\ddot{\theta}_1+M_{33}\ddot{\varphi}_1+\mu_g\dot{\theta}_1+(\mu_g+\mu_1)\dot{\varphi}_1=u_1,\tag{6}\label{eqn6}\\
&a_5\ddot{\theta}_2+M_{44}\ddot{\varphi}_2+\mu_g\dot{\theta}_2+\left(\mu_g+\mu_2\right)\dot{\varphi}_2=u_2.\tag{7}\label{eqn7}
\end{align*}
The parameters $a_1$, $a_2$, $a_3$, $a_4$, $a_5$, $a_6$ are shown in Appendix \ref{linear}. (\ref{eqn4})$-$(\ref{eqn7}) indicates that the linearized OmBURo dynamics can be completely decoupled in the $X$ and $Y$ directions. (\ref{eqn4}) and (\ref{eqn6}) are only related to $\theta_1$, $\varphi_1$, and $u_1$, while (\ref{eqn5}) and (\ref{eqn7}) are only related to $\theta_2$, $\varphi_2$, and $u_2$, which is essentially dual-axis wheeled inverted pendulum model. Solving (\ref{eqn4})$-$(\ref{eqn7}) leads to the state-space realization
\begin{align*}
\bm{\dot{x}}=\bm{Ax}+\bm{Bu},\tag{8}\label{eqn8}
\end{align*}
where
\begin{gather*}
{
\setlength{\arraycolsep}{4pt}
\medmath
{\bm{A}=\begin{bmatrix}
0&0&1&0&0&0\\
0&0&0&1&0&0\\
A_{31}&0&A_{33}&0&A_{35}&0\\
0&A_{42}&0&A_{44}&0&A_{46}\\
A_{51}&0&A_{53}&0&A_{55}&0\\
0&A_{62}&0&A_{64}&0&A_{66}\\
\end{bmatrix},~\bm{B}=\begin{bmatrix}
0&0\\
0&0\\
B_{31}&0\\
0&B_{42}\\
B_{51}&0\\
0&B_{62}\\
\end{bmatrix},~\bm{x}=\begin{bmatrix}
\theta_1\\
\theta_2\\
\dot{\theta}_1\\
\dot{\theta_2}\\
\dot{\varphi}_1\\
\dot{\varphi}_2
\end{bmatrix}}}.
\end{gather*}
The elements of $\bm{A}$ and $\bm{B}$ are listed in Appendix \ref{linear}. The pair of $(\bm{A},\bm{B})$ is verified to be controllable and all the states can be measured directly. 





\subsection{Locomotion Control}

\begin{figure}[!t]
    \centering
	\includegraphics[width=8cm]{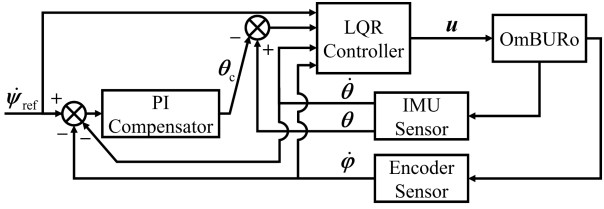}
	\caption {Locomotion control system}
	\label{fig: control block diagram}
\end{figure}

The objective of locomotion control for OmBURo is to track the reference velocity while keeping balance of its body. It is realized by the control system shown in Fig. \ref{fig: control block diagram}, which has one feedforward part and two closed loops: an inner loop control based on linear quadratic regulator (LQR) and an outer loop compensator using proportional-integral (PI) control. 

The feedforward part feeds the desired OmBURo velocity into the inner loop control as the reference signal. Note that $\bm{\dot{\psi}}_{\textrm{ref}}$ and $\bm{\dot{\varphi}}_{\textrm{ref}}$ are equivalent since $\bm{\dot{\psi}}=\bm{\dot{\varphi}}+\bm{\dot{\theta}}$ while $\bm{\dot{\theta}}=\bm{0}$ at steady state.

With full state feedback, LQR is used to track the constant reference velocity $\bm{\dot{\psi}}_{\textrm{ref}}$. Using the feedback control law $\bm{u} = -\bm{Kx} + \bm{H\dot{\psi}}_{\textrm{ref}}$, (\ref{eqn8}) becomes
\begin{align*}
\bm{\dot{x}}=&\left(\bm{A}-\bm{BK}\right)\bm{x}+\bm{BH}\bm{\dot{\psi}}_{\textrm{ref}}\\
\bm{y}=&\left[\begin{array}{cccccc}
\dot{\psi}_1\\
\dot{\psi}_2
\end{array}\right]=\underbrace{\left[\begin{array}{cccccc}
0&0&1&0&1&0\\
0&0&0&1&0&1
\end{array}\right]}_{\bm{C}}\bm{x}, \tag{9}
\end{align*}
where $\bm{K}$ is the gain matrix computed by MATLAB's \textit{lqr} command with proper choice of the weighting matrices. Considering the constant reference velocity $\bm{\dot{\psi}}_{\textrm{ref}}$ as a step input, the steady state output vector is thus given by
\begin{align*}
\lim_{t\to\infty}\bm{y}(t)=\bm{C}\left(\bm{BK}-\bm{A}\right)^{-1}\bm{BH\dot{\psi}}_{\textrm{ref}}. \tag{10}
\end{align*}
In order to track the reference well, $\bm{H}$ is determined to be
\begin{align*}
\bm{H}=\left(\bm{C}\left(\bm{BK}-\bm{A}\right)^{-1}\bm{B}\right)^{-1} \tag{11}
\end{align*}
so that $\displaystyle\lim_{t\to\infty}\bm{y}(t)=\bm{\dot{\psi}}_{\textrm{ref}}$. The inverse exists because there are as many control inputs as independent controlled outputs.

The outer loop control essentially compensates for the sensor bias as well as the modeling errors. Generally, it is inevitable to have these problems when implementing the control algorithms on the real physical system. For example, to keep OmBURo at a fixed position while balancing it, we have $\bm{\dot{\psi}}_{\textrm{ref}}=\bm{0}$ and (\ref{eqn8}) implies
\begin{align*}
\lim_{t\to\infty}\bm{\theta}(t)=-\mu_g\left[\begin{array}{cc}
1/a_3&0\\
0&1/a_6
\end{array}\right]\bm{\dot{\psi}}_{\textrm{ref}}=\left[\begin{array}{c}
0\\
0
\end{array}\right] \tag{12}\label{eqn12}
\end{align*}
by setting $\bm{\ddot{\theta}}=\bm{\ddot{\varphi}}=\bm{0}$ and  $\bm{\dot{\theta}}=\bm{0}$. However, the real balanced body angle is actually unknown and probably nonzero mainly due to IMU bias and shifted center of mass (CoM) from the central axis. Using just LQR, OmBURo will move towards its CoM and leave away from its initial position until it falls. With that in mind, we can actually use this information from the encoder to correct the body angle $\bm{\theta}$ via a PI compensator. The proportional and integral gains are experimentally tuned.

\section{Experimental Results}

Several tests were conducted to evaluate the performance of the physical system along with the proposed control strategy, which can be found online \href{https://youtu.be/avtlx1-X3lo}{(https://youtu.be/avtlx1-X3lo)}. Corresponding experimental results are demonstrated in this section. Note that in all the wheel trajectory plots, the coordinate system corresponds to Fig. \ref{fig: velocity kinematics} and OmBURo always starts at the origin.

\subsection{Balancing at a Fixed Position}

\begin{figure}[!t]
    \centering
    \includegraphics[width = 5.7cm]{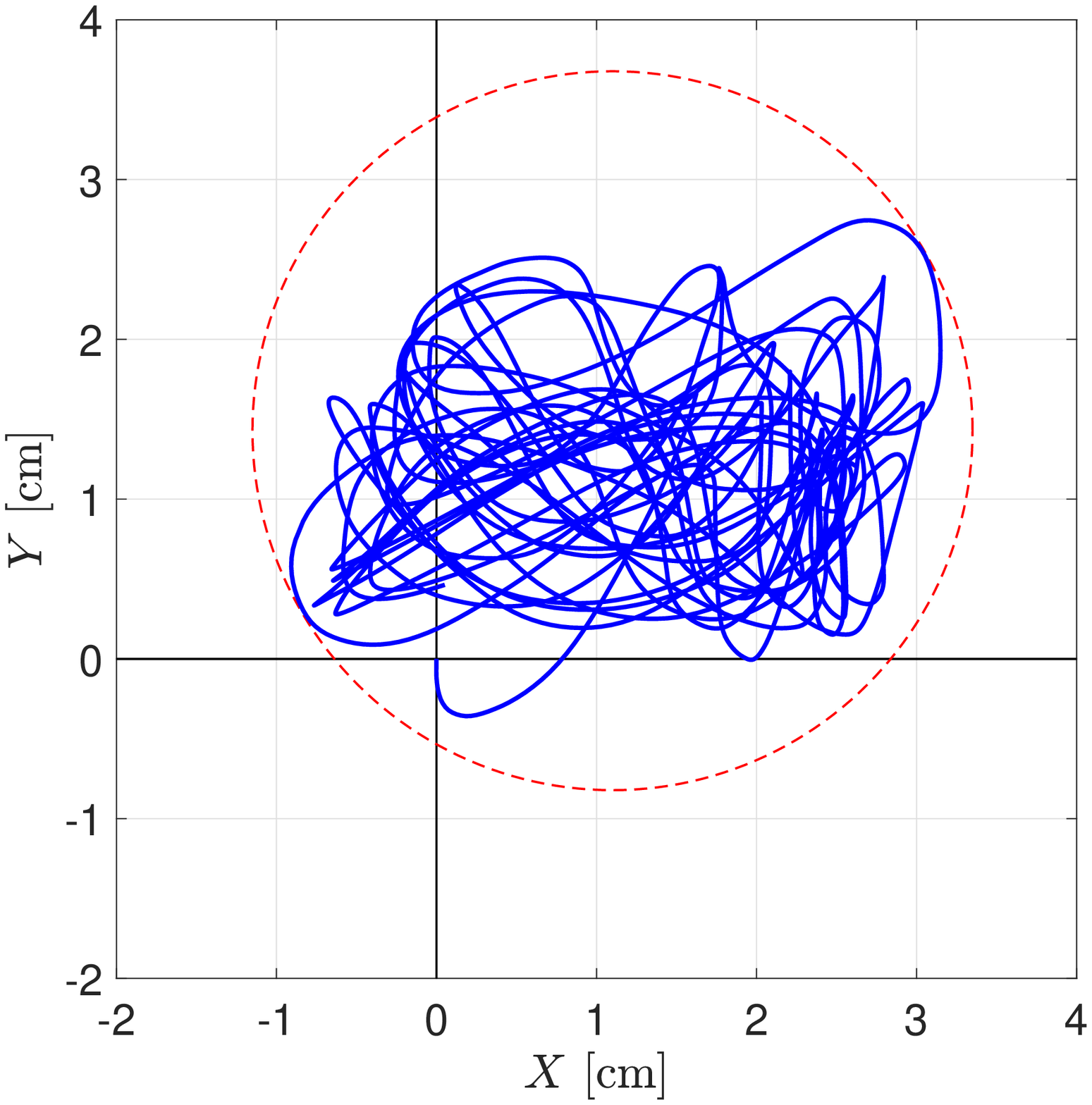}\\
    \scriptsize{
    (a) Wheel trajectory
    }
    \includegraphics[height = 4.2cm]{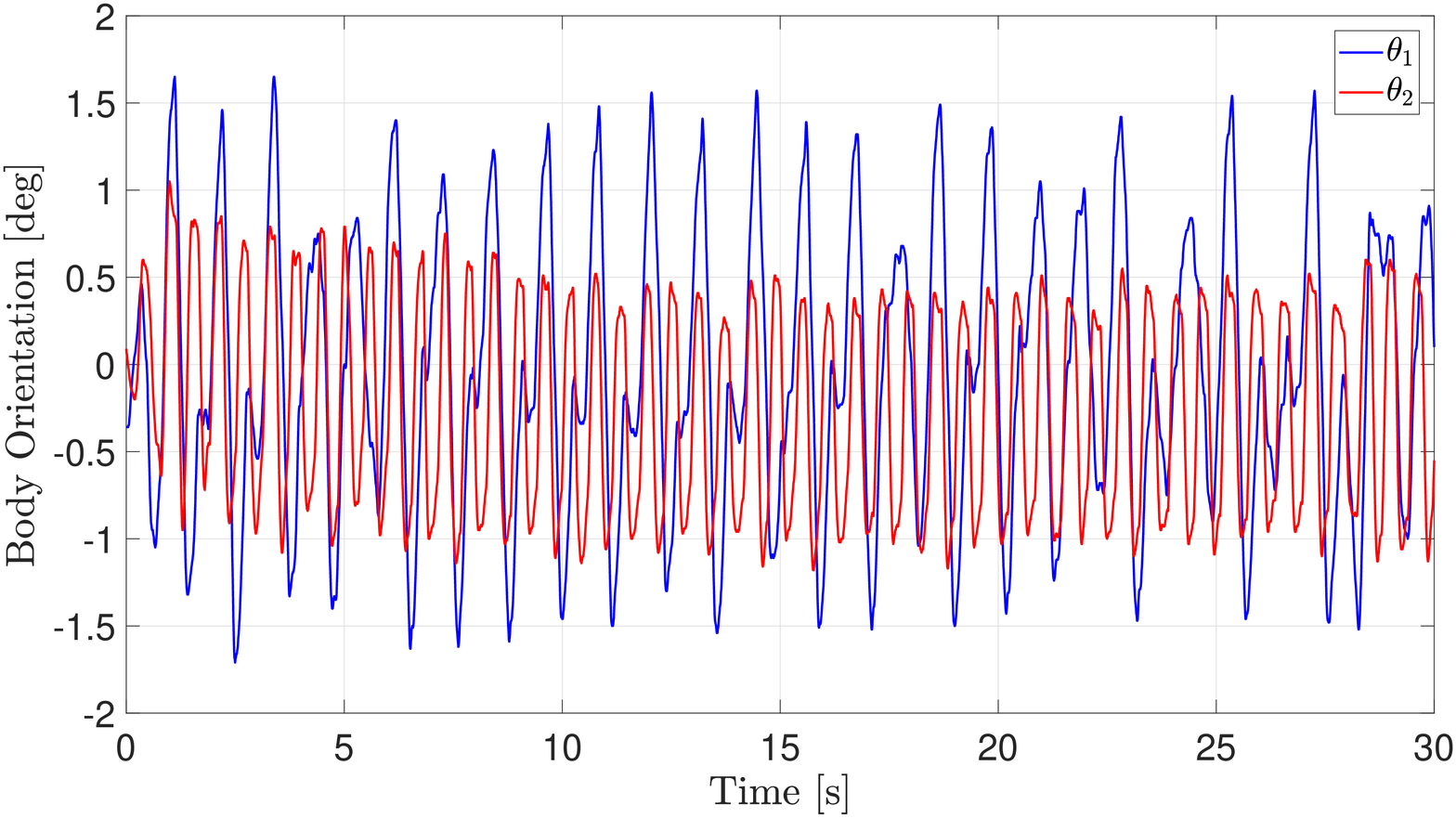}\\
    \scriptsize{(b) Time series of body orientation}
    \caption{Experimental results of balancing at a fixed position. (a) At steady state, OmBURo was unexpectedly taking an $\infty$-shaped wheel trajectory (blue line) within a circular region (red dashed line) with a diameter of about 4.5 cm. (b) Both $\theta_1$ and $\theta_2$ were oscillating around nonzero values due to IMU bias and shifted CoM. $\theta_2$ has twice the frequency as $\theta_1$, which corresponds to the $\infty$-shaped wheel trajectory.} 
    \label{fig: balancing}
\end{figure}

In this test, we want to keep OmBURo at a fixed position while balancing it. Accordingly, we have $\bm{\dot{\psi}}_{\textrm{ref}}=\bm{0}$. OmBURo was released out of balance but was able to quickly correct its body orientation via its wheel motion. Fig. \ref{fig: balancing}(a) shows its wheel trajectory. We can see that OmBURo was keeping its position within the neighborhood of some point at steady state, a nearly circular region (red dashed line) with a diameter of about 4.5 cm. Fig. \ref{fig: balancing}(b) shows the corresponding time series of its body orientation. Note that the real balanced body angle (from IMU) is nonzero, as mentioned before.

\begin{figure}[!t]
    \centering
    \includegraphics[width = 5.7cm]{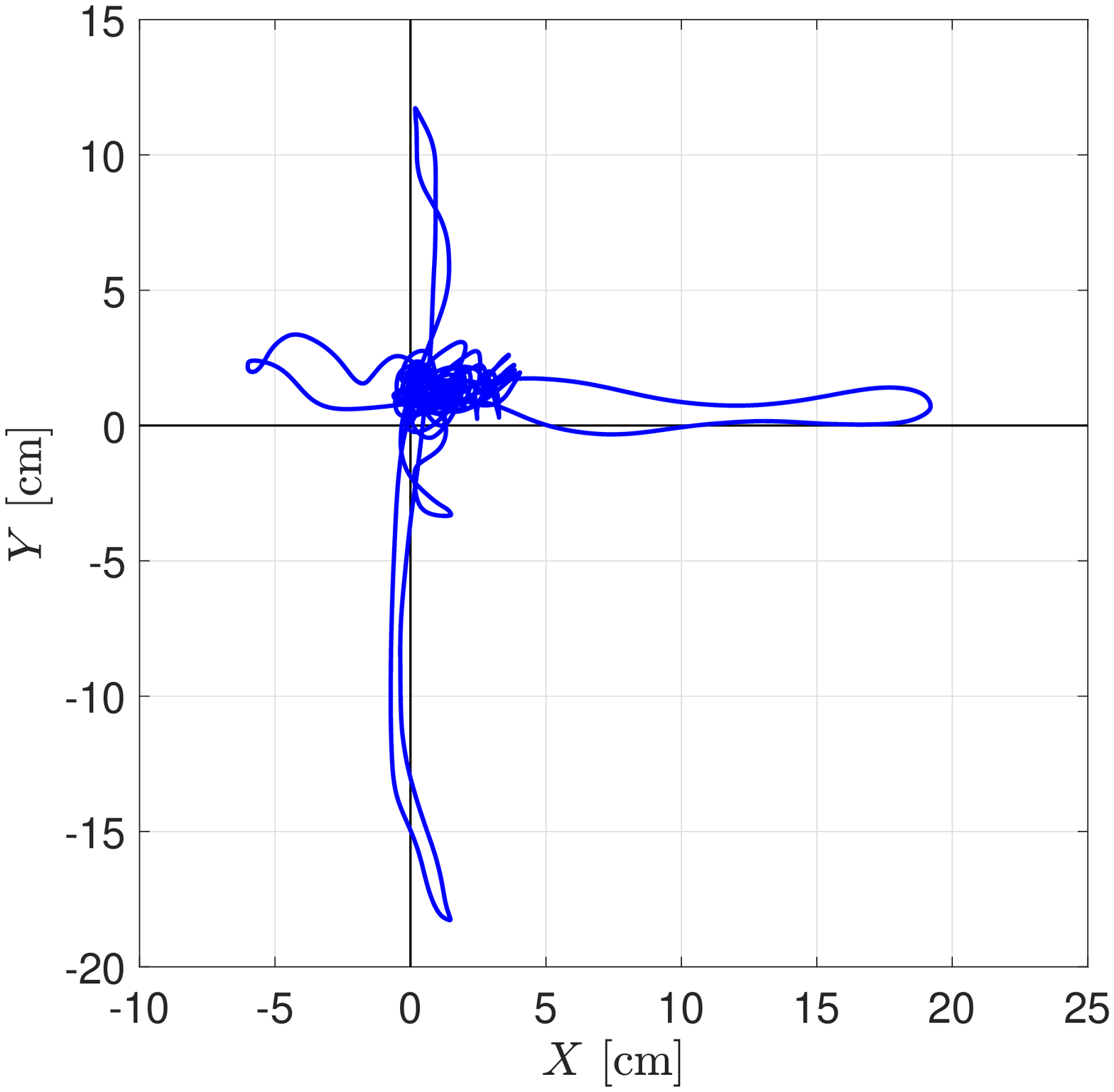}\\
    \scriptsize{(a) Wheel trajectory}\\
    \includegraphics[height = 4.2cm]{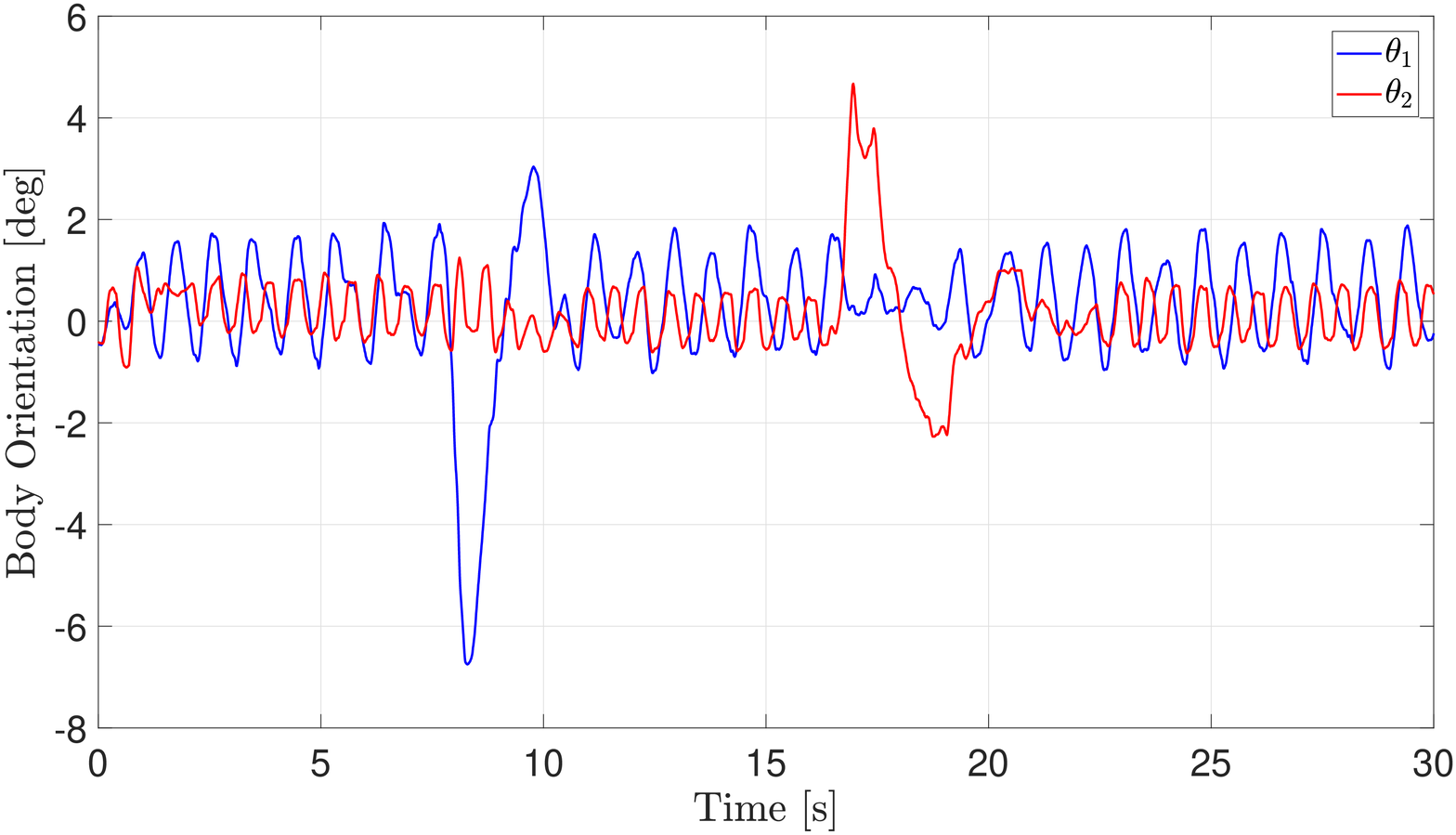}\\
    \scriptsize{(b) Time series of body orientation}
    \caption{Experimental results of disturbance rejection. While station keeping, OmBURo was intentionally pushed in the positive $X-$direction around $t=8$ s and later in the $Y-$direction around $t=17$ s. It was able to recover within 3 seconds in both scenarios.}
    \label{fig: disturbance}
\end{figure}

\subsection{Disturbance Rejection}

To test the robustness of the whole system, a disturbance rejection experiment was carried out based on station keeping. When OmBURo was balancing itself at a fixed position, its body was intentionally pushed in the positive $X-$direction at time around 8 seconds. It regained its balance and went back to the original position within 3 seconds, as shown in Fig. \ref{fig: disturbance}. Another push was made in the negative $Y-$direction at time around 17 seconds and OmBURo was able to recover as well. The large deviation from the initial condition implies a quite decent stability region for the prototype and it can be further improved by using more powerful actuators.

\subsection{Velocity Tracking}

\begin{figure}[!t]
    \centering
    \includegraphics[height = 4.2cm]{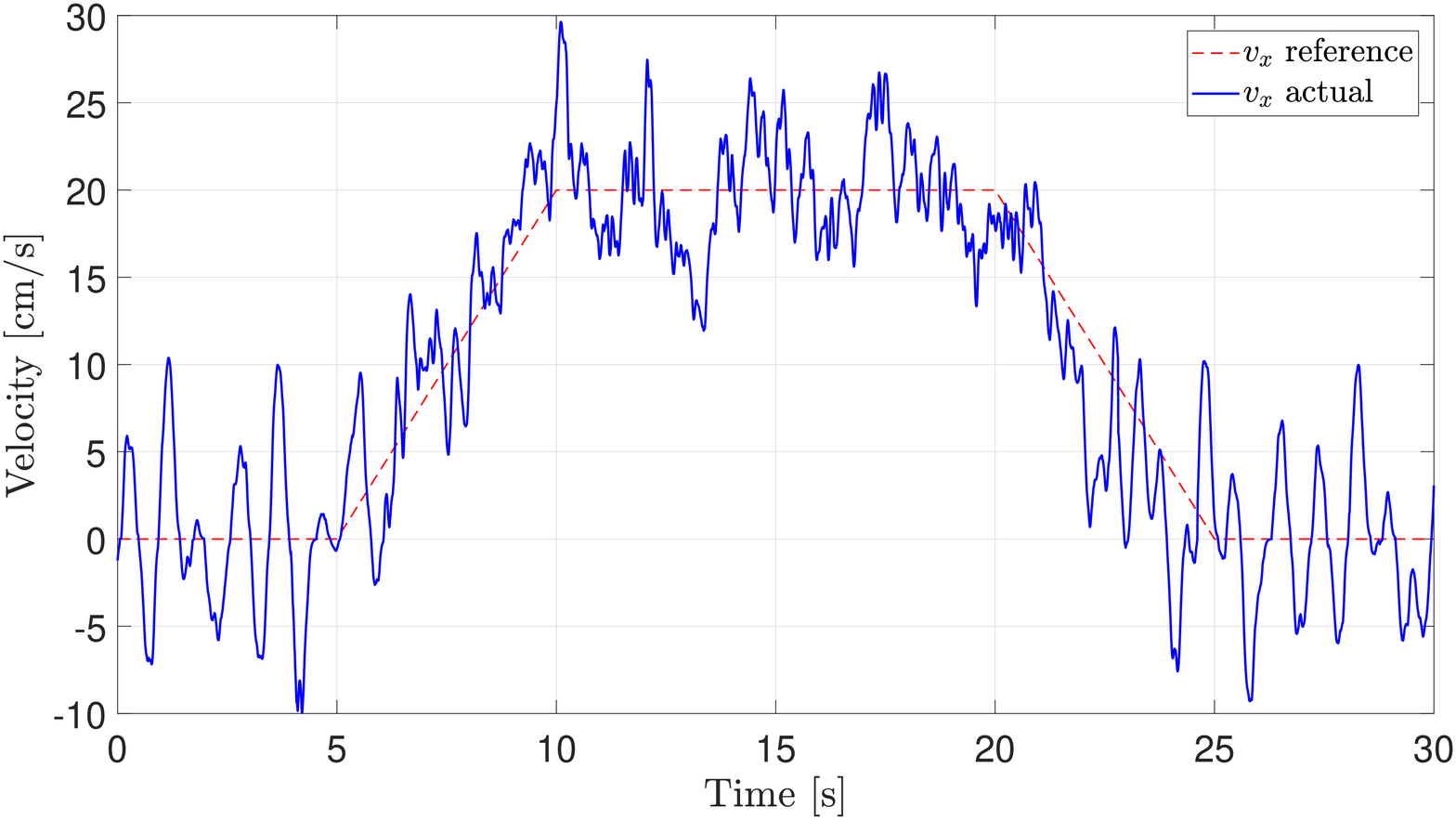}\\
    \scriptsize{(a) Time series of wheel velocity}\\
    \includegraphics[height = 4.2cm]{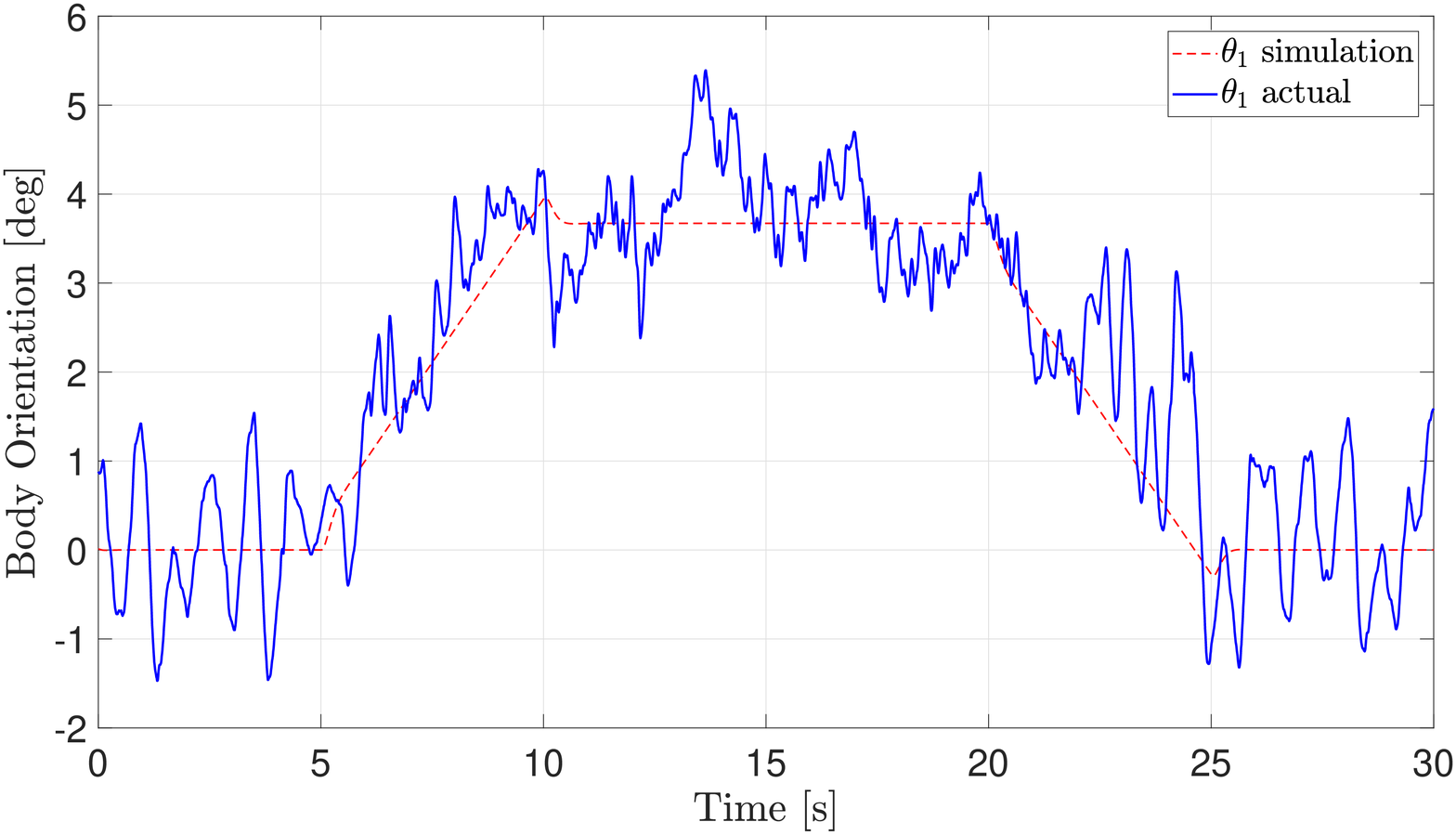}\\
    \scriptsize{(b) Time series of body orientation}
    \caption{Experimental results of velocity tracking. In order for OmBURo to move with a constant velocity, its body needs to lean for a certain angle against ground friction, as suggested by (\ref{eqn12}). Due to limited space, results are shown only for the $X-$direction.} 
    \label{fig: velocity tracking}
\end{figure}

In this test, we want to move OmBURo with a constant velocity. To avoid latent instability caused by the sudden state change, a trapezoid reference is predefined instead of a step reference. Fig. \ref{fig: velocity tracking} shows data compared with the reference. One interesting observation is that the wheel motion almost always had an opposite behavior to body orientation. The fact is that to move in some given direction, OmBURo however starts with a quick opposite movement, causing its body to lean towards the goal direction due to inertia, and then reverses its wheel and speeds up using its dynamics. 

\subsection{Path Following}

\begin{figure}[!t]
    \centering
    \includegraphics[width = 5.7cm]{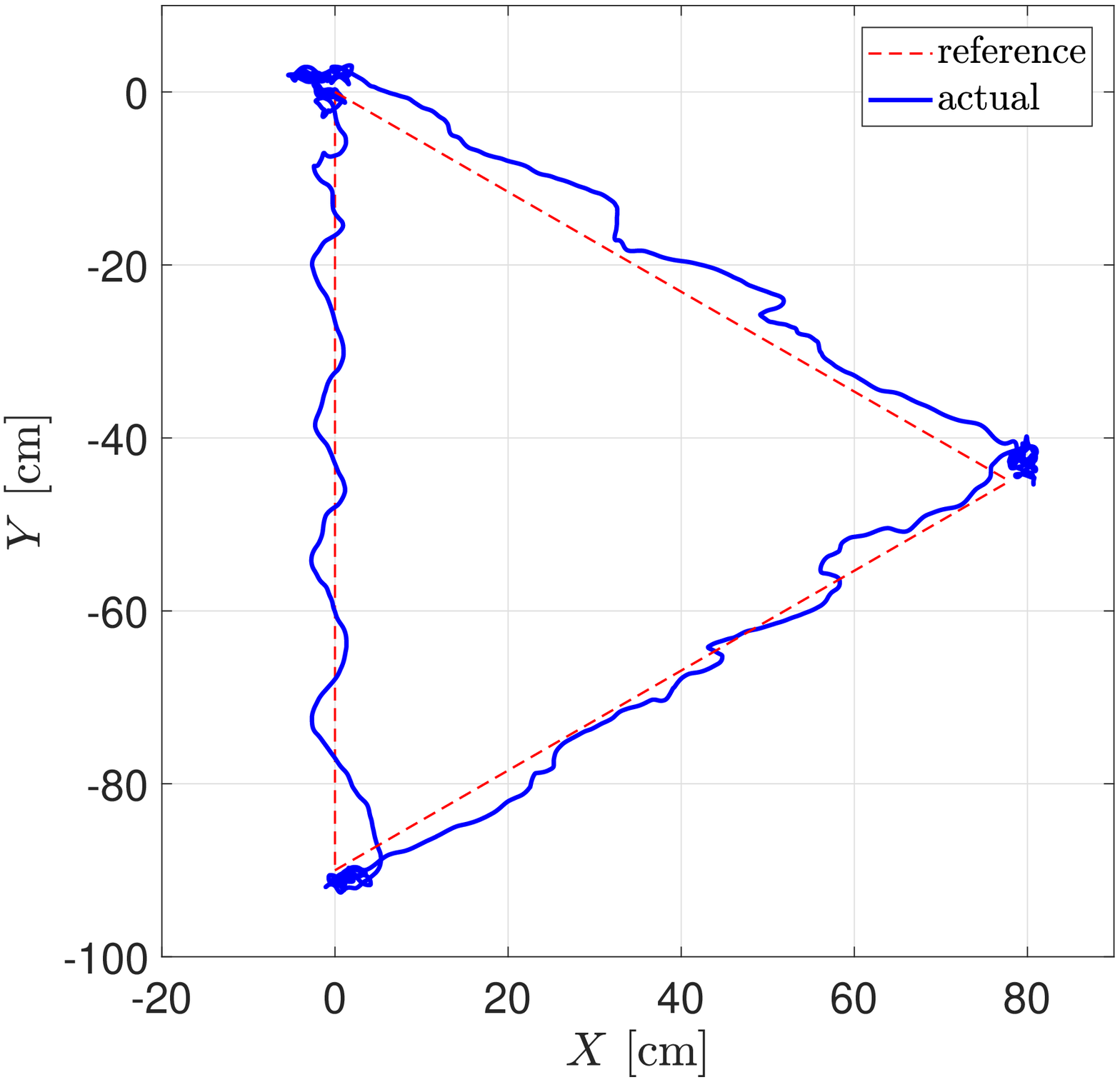}\\
    \scriptsize{(a) A triangular path}\\
    \includegraphics[width = 5.7cm]{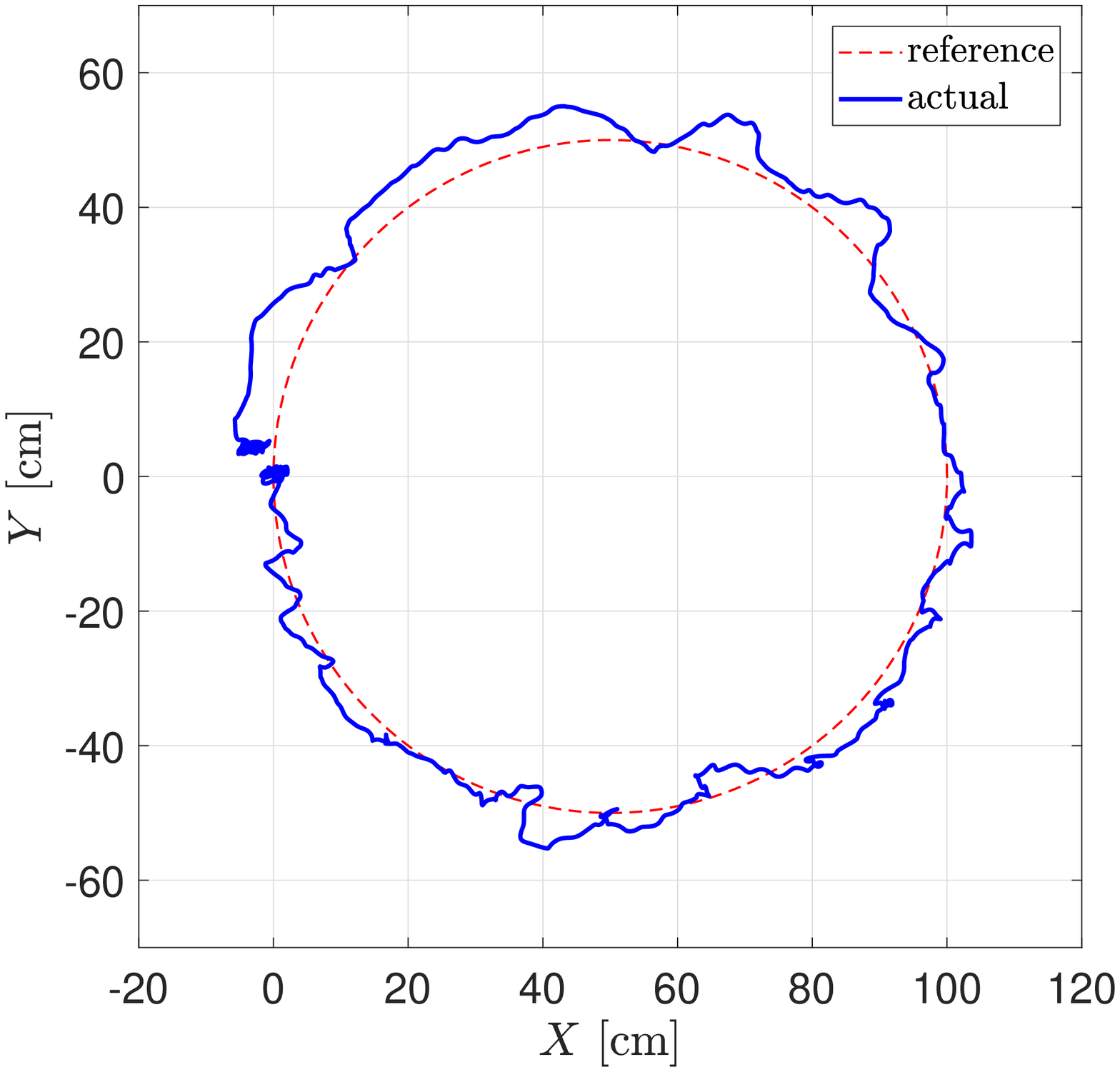}\\
    \scriptsize{(b) A circular path}
    \caption{Experimental results of path following. The red dashed line is the reference while the blue line is the actual wheel trajectory. OmBURo was remotely controlled to adjust its velocity reference during the test.} 
    \label{fig: path following}
\end{figure}

In this test, we want to make OmBURo follow some predefined paths. Since it has no extrinsic sensors for now, OmBURo was remotely controlled to adjust its velocity reference during the test. Fig. \ref{fig: path following} shows two scenarios, namely  (a) a triangular path and (b) a circular path. These motions were intentionally made slow and steady so that OmBURo could be tightly controlled to follow the path well, but it might not be the best approach judging from the jagged profile. Nevertheless, the capability of following these paths indicates that the overall platform is good enough to implement trajectory tracking in terms of both desired position and velocity.

\begin{figure}[!t]
    \centering
    \includegraphics[width = 5.7cm]{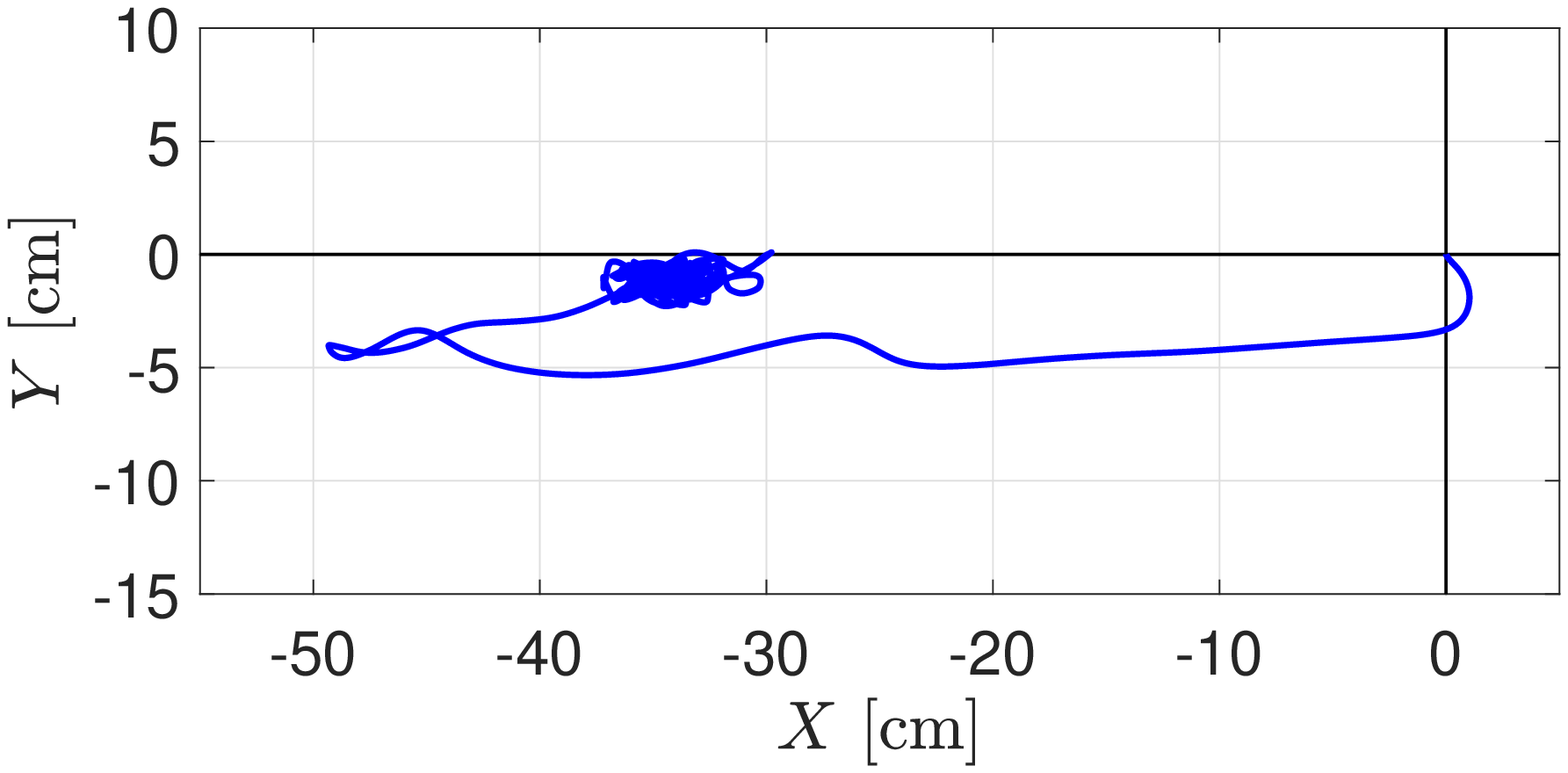}\\
    \scriptsize{(a) Wheel trajectory $\left(\gamma=15^{\circ}\right)$}\\
    \includegraphics[height = 4.2cm]{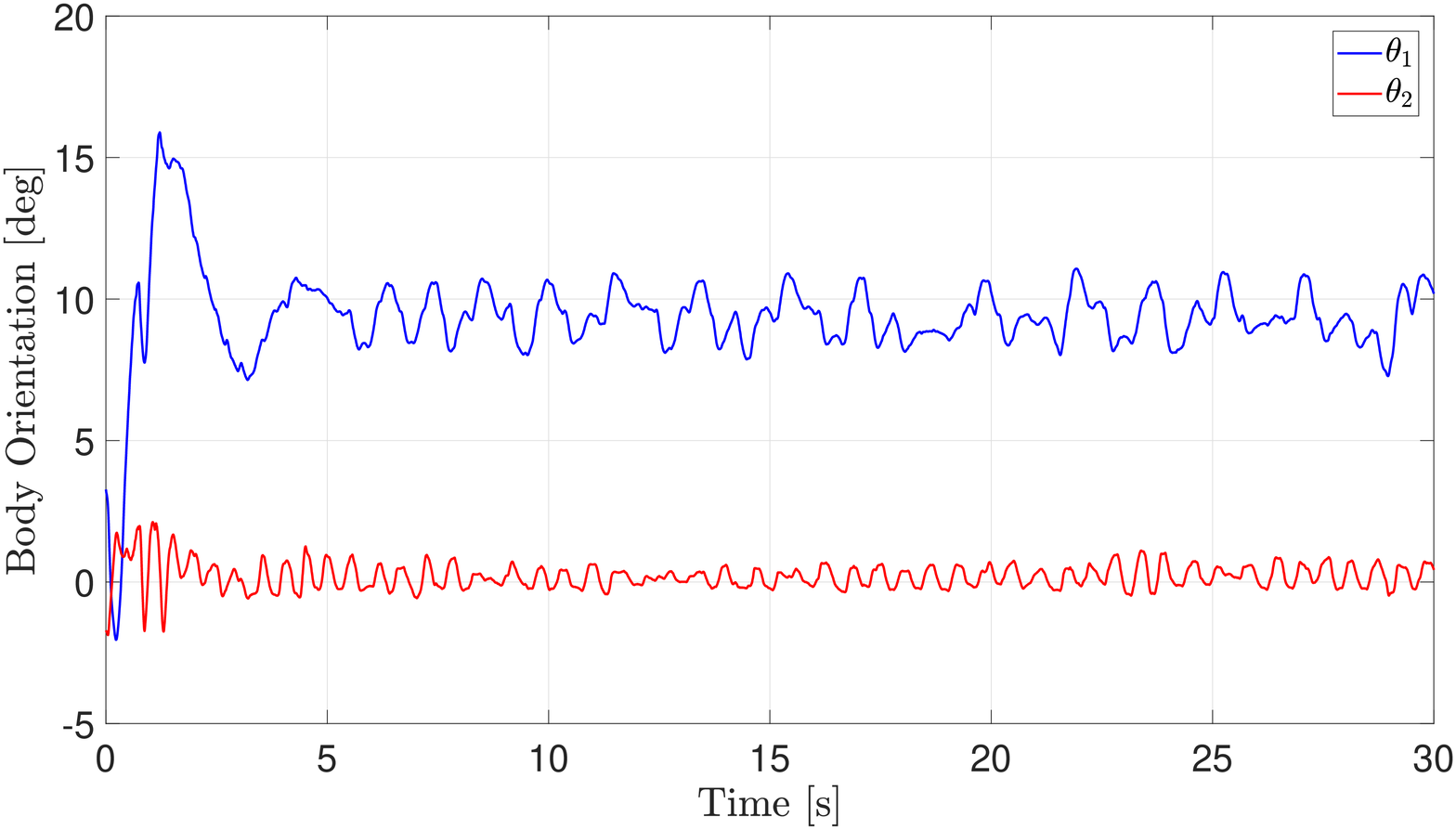}\\
    \scriptsize{(b) Time series of body orientation $\left(\gamma=15^{\circ}\right)$}\\
    \includegraphics[height = 3.9cm]{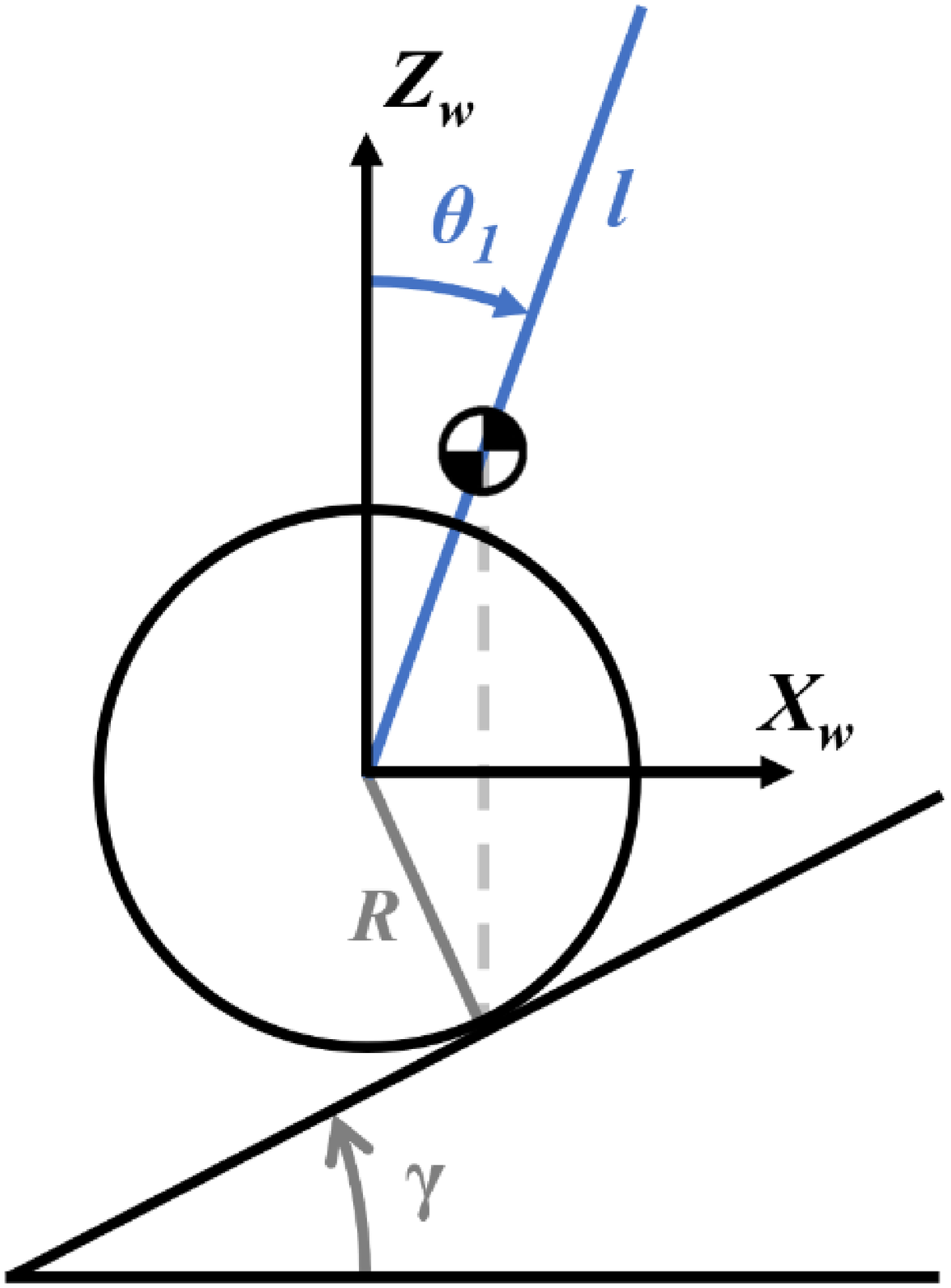}$~~~$
    \includegraphics[height = 4.3cm]{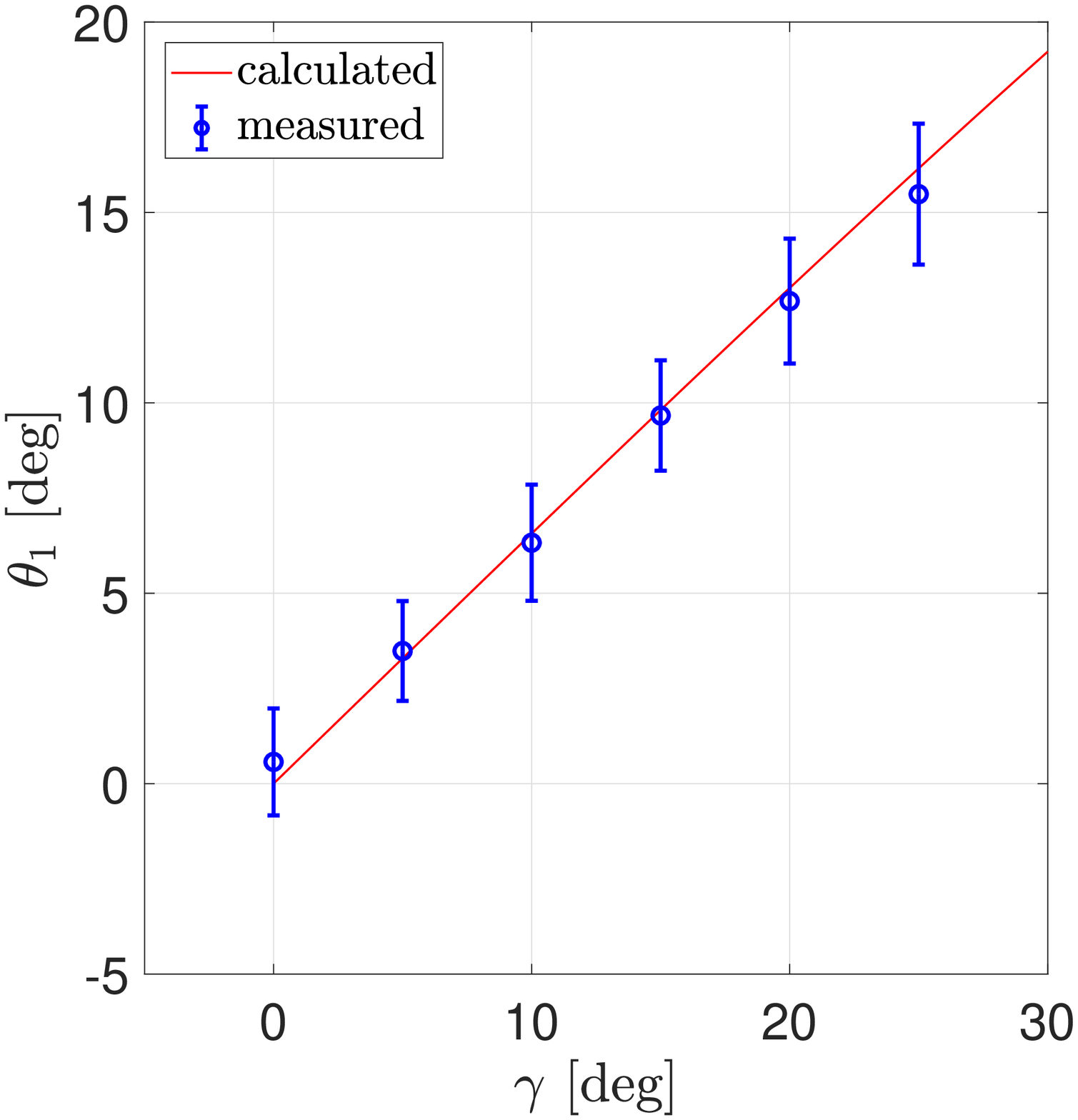}\\
    \scriptsize{(c) OmBURo aligned with inclination $~~~~~~~~~~~~~~~$(d) $\theta_1$ vs. $\gamma~~~~~~~~~~~~~~~~~~~~~~~~$}
    \caption{Experimental results of balancing on an inclined plane. After release, OmBURo first rolled down the slope for some distance but was still capable of station keeping eventually. Furthermore, it was able to adapt to different inclination angles by adjusting its body orientation.} 
    \label{fig: inclined}
\end{figure}

\subsection{Balancing on an Inclined Plane}

Finally, we tried to make OmBURo balance itself on an inclined plane. Similar to balancing on a horizontal plane, we set $\bm{\dot{\psi}}_{\textrm{ref}}=\bm{0}$ and just released it. It first rolled down the slope for some distance but was still able to quickly balance itself and eventually keeping its position within some area, as shown in Fig. \ref{fig: inclined}(a, b). Ideally, in order to balance on a slope, its CoM should be right above the contact point. For simplicity, here we are only considering the situation when OmBURo is aligned with the inclination angle, as shown in Fig. \ref{fig: inclined}(c). The relationship between the pitch angle $\theta_1$ and the inclination angle $\gamma$ is thus determined to be
\begin{align*}
m_bl\sin\theta_1=\left(m_b+m_w\right)R\sin\gamma.\tag{13}
\end{align*}
Fig. \ref{fig: inclined}(d) shows calculated values compared with measured data range at steady state. The largest feasible inclination angle is around 30 degrees, presumably limited by the ground friction coefficient. Unlike Segway type robots, which might fall over on an inclined plane if not aligned with its moving direction, we further changed the yaw angle manually during the test and OmBURo was still capable of station keeping. 

\section{Conclusion}

In this letter, first, a novel active omnidirectional wheel was proposed. For conventional omni wheels, the peripheral rollers can only be passively driven. As a result, several omni wheels have to work together for omnidirectional drive system. Since the rollers can now be directly actuated, a single wheel unit is sufficient to actively move in any arbitrary direction. 

Second, a comparison with the existing omnidirectional mobility mechanisms was conducted. Friction drive transmission on Ballbot and U3-X is simple and elegant. However, it has limitations on ill-conditioned terrains. It also suffers rapid tire wear and energy loss as heat. Gear transmission avoids these problems but results in a large gap between the rollers. Future work will be concentrated on how to minimize the gap for a smoother rolling motion. Besides, a better structure can be investigated to drive the bottom roller only while freeing the rest, thus enhancing energy efficiency.

Third, a new unicycle robot, OmBURo, was built based on the proposed active omnidirectional wheel. Accordingly, an interesting unicycle model was derived and a corresponding locomotion control strategy was designed. The performance is good enough to implement trajectory tracking in terms of both desired position and velocity. It might be eventually used in human environments due to its very simple and compact structure, as well as agile and omnidirectional mobility. Future work will be further focused on its locomotion control design, e.g., steering control and trajectory tracking. 

\appendix
\subsection{Derivation of Dynamics}
\label{dynamics}
From Fig. \ref{fig: model}, the positions of the roller frame, wheel frame, and body frame in the inertial frame are determined to be
\setlength{\abovedisplayskip}{2.9pt}
\setlength{\belowdisplayskip}{2.9pt}
{\footnotesize
\begin{align*}
\left[\begin{array}{cc}
x_r\\
y_r\\
z_r
\end{array}\right]&=\left[\begin{array}{cc}
R\psi_1\\
r\psi_2\\
r
\end{array}\right],\quad\left[\begin{array}{cc}
x_w\\
y_w\\
z_w
\end{array}\right]=\left[\begin{array}{cc}
x_r\\
y_r+\left(R-r\right)\sin\theta_2\\
z_r+\left(R-r\right)\cos\theta_2
\end{array}\right],
\\
\left[\begin{array}{cc}
x_b\\
y_b\\
z_b
\end{array}\right]&=\left[\begin{array}{cc}
x_w+l\sin\theta_1\\
y_w+l\cos\theta_1\sin\theta_2\\
z_w+l\cos\theta_1\cos\theta_2
\end{array}\right].
\end{align*}
}\noindent
\noindent
Therefore, the total kinetic energy is determined to be
{\footnotesize
$$
K=K_r+K_w+K_b,
$$
}\noindent
where
\allowdisplaybreaks
{\footnotesize
\begin{align*}
K_r=&~\frac{1}{2}I_{rx}\dot{\psi}_2^2,\\
K_w=&~\frac{1}{2}\left[\begin{array}{cc}
\dot{\theta}_2\cos\psi_1\\
\dot{\psi}_1\\
\dot{\theta}_2\sin\psi_1
\end{array}\right]^T\underbrace{\left[\begin{array}{ccc}
I_{wx}&0&0\\
0&I_{wy}&0\\
0&0&I_{wz}\\
\end{array}\right]}_{\bm{I_w}}\left[\begin{array}{cc}
\dot{\theta}_2\cos\psi_1\\
\dot{\psi}_1\\
\dot{\theta}_2\sin\psi_1
\end{array}\right]\\
&+\frac{1}{2}m_w\left(\dot{x}_w^2+\dot{y}_w^2+\dot{z}_w^2\right),~\left(I_{wx}=I_{wz}\textrm{ due to symmetry}\right)\\
K_b=&~\frac{1}{2}\left[\begin{array}{cc}
\dot{\theta}_2\cos\theta_1\\
\dot{\theta}_1\\
\dot{\theta}_2\sin\theta_1
\end{array}\right]^T\underbrace{\left[\begin{array}{ccc}
I_{bx}&0&0\\
0&I_{by}&0\\
0&0&I_{bz}\\
\end{array}\right]}_{\bm{I_b}}\left[\begin{array}{cc}
\dot{\theta}_2\cos\theta_1\\
\dot{\theta}_1\\
\dot{\theta}_2\sin\theta_1
\end{array}\right]\\
&+\frac{1}{2}m_b\left(\dot{x}_b^2+\dot{y}_b^2+\dot{z}_b^2\right),
\end{align*}
}\noindent
and $m_w$, $m_b$ are the masses of the omni wheel and the body, $\bm{I_w}$ and $\bm{I_b}$ are the inertia tensors of the omni wheel and the body about their principal axes, and $I_{rx}$ is the moment of inertia of all the rollers about each central axis. On the assumption of rigid bodies, the total potential energy of OmBURo is determined to be
{\footnotesize
$$
U=U_w+U_b=m_wgz_w+m_bgz_b
$$
}\noindent
due only to gravitational effect. The dissipation energy is determined to be
{\footnotesize
$$
D=\frac{1}{2}\mu_g\left(\dot{\psi}_1^2+\dot{\psi}_2^2\right)+\frac{1}{2}\mu_1\dot{\varphi}_1^2+\frac{1}{2}\mu_2\dot{\varphi}_2^2,
$$
}\noindent
where $\mu_g$ is the viscous coefficient between the roller and the ground while $\mu_1$ and $\mu_2$ are the viscous coefficients of the wheel axle and the roller axle respectively. The Lagrangian is determined by
{\footnotesize
$$
L=K-U
$$
}\noindent
and the Lagrange equation is derived by
{\footnotesize
$$
\frac{d}{dt}\left(\frac{\partial L}{\partial\bm{\dot{q}}}\right)-\frac{\partial L}{\partial\bm{q}}+\frac{\partial D}{\partial\bm{\dot{q}}}=\frac{\partial\bm{\varphi}}{\partial\bm{q}}\bm{u}.
$$
}\noindent
After calculating the derivatives, rearranging and simplifying terms, the nonlinear dynamics is governed by
{\setlength{\arraycolsep}{1.2pt}
{\footnotesize
\begin{align*}
\left[\begin{array}{cccc}
M_{11}&M_{12}&M_{13}&M_{14}\\
M_{21}&M_{22}&0&M_{24}\\
M_{31}&0&M_{33}&0\\
M_{41}&M_{42}&0&M_{44}
\end{array}\right]\left[\begin{array}{cccc}
\ddot{\theta}_1\\
\ddot{\theta}_2\\
\ddot{\varphi}_1\\
\ddot{\varphi}_2
\end{array}\right]+\left[\begin{array}{cccc}
C_{1}\\
C_{2}\\
C_{3}\\
C_{4}
\end{array}\right]+\left[\begin{array}{cccc}
F_1\\
F_2\\
F_3\\
F_4
\end{array}\right]+\left[\begin{array}{cccc}
G_1\\
G_2\\
0\\
0
\end{array}\right]=\left[\begin{array}{cccc}
0\\
0\\
u_1\\
u_2
\end{array}\right],
\end{align*}
}
}\noindent
\noindent
where
\allowdisplaybreaks
{\footnotesize
\begin{align*}
M_{11}=&~I_{by}+I_{wy}+m_bl^2+\underbrace{\left(m_b+m_w\right)}_{m}R^2+2m_blR\cos\theta_1,\\
M_{12}=&~M_{21}=-m_blr\sin\theta_1\sin\theta_2,\\
M_{13}=&~M_{31}=I_{wy}+mR^2+m_blR\cos\theta_1,\\
M_{14}=&~M_{41}=-m_blr\sin\theta_1\sin\theta_2,\\
M_{22}=&~\frac{1}{2}\left(I_{bx}+I_{bz}+m_bl^2\right)+I_{rx}+I_{wx}+m\left(2 r^2-2 r R+R^2\right)\\
&~+2mr(R-r) \cos\theta_2+\frac{1}{2}\left(I_{bx}-I_{bz}+m_bl^2 \right)\cos2\theta_1\\
&~+2m_bl\cos\theta_1\underbrace{\left(R+r\cos\theta_2-r\right)}_{d},\\
M_{24}=&~M_{42}=I_{rx}+mr^2+r\left[m(R-r)+m_bl\cos\theta_1\right]\cos\theta_2,\\
M_{33}=&~I_{wy}+mR^2,\quad M_{44}=I_{rx}+mr^2,\\
C_1=&~\left[\left(I_{bx}-I_{bz}+m_bl^2\right)\cos\theta_1+m_bl(R-r)\right]\dot{\theta}_2^2\sin\theta_1\\
&~-m_blR\dot{\theta}_1^2\sin\theta_1,\\
C_2=&~2\left[\left(I_{bz}-I_{bx}-m_bl^2\right)\cos\theta_1-m_bld\right]\dot{\theta}_1\dot{\theta}_2\sin\theta_1\\
&~-\left[m(R-r)+m_bl\cos\theta_1\right]r\dot{\theta}_2^2\sin\theta_2\\
&~-m_blr\dot{\theta}_1^2\cos\theta_1\sin\theta_2,\\
C_3=&-m_blR\dot{\theta}_1^2\sin\theta_1,\\
C_4=&-m_blr\dot{\theta}_1^2\cos\theta_1\sin\theta_2-2m_blr\dot{\theta}_1\dot{\theta}_2\sin\theta_1\cos\theta_2\\
&-\left[m(R-r)+m_bl\cos\theta_1\right]r\dot{\theta}_2^2\sin\theta_2,\\
F_1=&~\mu_g(\dot{\theta}_1+\dot{\varphi}_1),\quad ~~~~~~~~F_2=\mu_g(\dot{\theta}_2+\dot{\varphi}_2),\\
F_3=&~\mu_g(\dot{\theta}_1+\dot{\varphi}_1)+\mu_1\dot{\varphi}_1,~~F_4=\mu_g(\dot{\theta}_2+\dot{\varphi}_2)+\mu_2\dot{\varphi}_2,\\
G_1=&-m_bgl\sin\theta_1\cos\theta_2,~~~G_2=\left[m(R-r)+m_bl\cos\theta_1\right]g\sin\theta_2.
\end{align*}
}\noindent
All the parameters and variables are listed in Table \ref{tab: parameters}.

\begin{table}[ht]
\caption{OmBURo Parameters and Variables}
\begin{center}
\begin{tabular}{c|l|l}
\hline
\bf{Symbol}&\bf{Parameter \& Variable}&\bf{Value \& Unit}\\ \hline
$R$&Radius of the wheel&0.101 m\\
$r$&Radius of the roller&0.0142 m\\
$l$&Distance between the wheel and body&0.2511 m\\
$m_w$&Mass of the wheel&0.72 kg\\
$m_b$&Mass of the body&1.13 kg\\
$I_{wx}$&Moment of inertia of the wheel in $X_w$&1.925 g$\cdot$m$^2$\\
$I_{wy}$&Moment of inertia of the wheel in $Y_w$&3.706 g$\cdot$m$^2$\\
$I_{wz}$&Moment of inertia of the wheel in $Z_w$&1.925 g$\cdot$m$^2$\\
$I_{bx}$&Moment of inertia of the body in $X_b$&38.909 g$\cdot$m$^2$\\
$I_{by}$&Moment of inertia of the body in $Y_b$&2.608 g$\cdot$m$^2$\\
$I_{bz}$&Moment of inertia of the body in $Z_b$&38.256 g$\cdot$m$^2$\\
$I_{rx}$&Moment of inertia of the roller in $X_r$&0.030 g$\cdot$m$^2$\\
$\mu_g$&Viscous between the wheel and ground&9 N$\cdot$cm/(rad/s)\\
$\mu_1$&Viscous of the wheel axle&0.5 N$\cdot$cm/(rad/s)\\
$\mu_2$&Viscous of the roller axle&0.5 N$\cdot$cm/(rad/s)\\
$g$&Gravitational acceleration&9.81 m/s$^2$\\
\hline
$\theta_1/\theta_2$&Pitch/roll angle&rad\\
$\psi_1/\psi_2$&Rotation angle of the wheel/roller&rad\\
$\varphi_1/\varphi_2$&Relative angle of wheel/roller to body&rad\\
$u_1$, $u_2$&Generalized forces&N$\cdot$m\\
\hline
\end{tabular}
\label{tab: parameters}
\end{center}
\end{table}

\addtolength{\textheight}{-0.84cm}   


\subsection{Parameters in State-space Realization}
\label{linear}
{\footnotesize
\begin{align*}
a_1=&~I_{by}+I_{wy}+m_bl^2+mR^2+2m_blR,\\
a_2=&~I_{wy}+mR^2+m_blR,\quad a_3=-m_bgl,\\
a_4=&~I_{bx}+I_{rx}+I_{wx}+m_bl^2+mR^2+2m_blR,~~~~~~~~~~~~~~~~~~~~~~\\
a_5=&~I_{rx}+mrR+m_blr,\quad a_6=-m_bgl-mg(R-r),
\end{align*}
\begin{align*}
A_{31}=&~\frac{a_3M_{33}}{a_2^2-a_1M_{33}}, &&A_{33}=\frac{\mu_g\left(M_{33}-a_2\right)}{a_2^2-a_1M_{33}},\\
A_{35}=&~\frac{\mu_gM_{33}-\left(\mu_g+\mu_1\right)a_2}{a_2^2-a_1M_{33}}, &&A_{42}=\frac{a_6M_{44}}{a_5^2-a_4M_{44}},\\
A_{44}=&~\frac{\mu_g\left(M_{44}-a_5\right)}{a_5^2-a_4M_{44}},&& A_{46}=\frac{\mu_gM_{44}-\left(\mu_g+\mu_2\right)a_5}{a_5^2-a_4M_{44}},\\
A_{51}=&~\frac{-a_2a_3}{a_2^2-a_1M_{33}},&& A_{53}=\frac{\mu_g\left(a_1-a_2\right)}{a_2^2-a_1M_{33}},\\
A_{55}=&~\frac{-\mu_ga_2+\left(\mu_g+\mu_1\right)a_1}{a_2^2-a_1M_{33}},&& A_{62}=\frac{-a_5a_6}{a_5^2-a_4M_{44}},\\
A_{64}=&~\frac{\mu_g\left(a_4-a_5\right)}{a_5^2-a_4M_{44}},&& A_{66}=\frac{-\mu_ga_5+\left(\mu_g+\mu_2\right)a_4}{a_5^2-a_4M_{44}},\\
B_{31}=&~\frac{a_2}{a_2^2-a_1M_{33}},&& B_{42}=\frac{a_5}{a_5^2-a_4M_{44}},\\
B_{51}=&~\frac{-a_1}{a_2^2-a_1M_{33}},&& B_{62}=\frac{-a_4}{a_5^2-a_4M_{44}}.
\end{align*}
}

\section*{Acknowledgment}

This work is partially supported by ONR through grant N00014-15-1-2064. The authors thank the Robotics and Mechanisms Laboratory at UCLA. In particular, the authors thank Xiaoguang Zhang and Yeting Liu for helping realize the physical system as well as Zhaoxing Deng for helping with the experiments.



{
\bibliographystyle{ieeetr}
\bibliography{references}
}

\end{document}